\documentclass{article}
\usepackage{arxiv}

\usepackage[utf8]{inputenc} 
\usepackage[T1]{fontenc}    
\usepackage{hyperref}       
\usepackage{url}            
\usepackage{booktabs}       
\usepackage{amsfonts}       
\usepackage{nicefrac}       
\usepackage{microtype}      
\usepackage{lipsum}		
\usepackage{graphicx}
\usepackage{natbib}
\usepackage{doi}
\usepackage{subfigure}
\usepackage{amsmath}
\usepackage{algpseudocode}
\usepackage{algorithm2e}

\title{Reinforcement Learning for Safe Occupancy Strategies in Educational Spaces during an Epidemic}
\date{}

\author{ {Elizabeth Akinyi Ondula}\\
	Viterbi School of Engineering\\
	University of Southern California\\
	Los Angeles, CA, USA \\
	\texttt{ondula@usc.edu} \\
	\And
	{Bhaskar Krishnamachari}\\
	Viterbi School of Engineering\\
	University of Southern California\\
	Los Angeles, CA, USA \\
	\texttt{bkrishna@usc.edu} \\
}

\renewcommand{\shorttitle}{RL for Safe Occupancy Strategies}

\hypersetup{
pdftitle={A template for the arxiv style},
pdfsubject={q-bio.NC, q-bio.QM},
pdfauthor={David S.~Hippocampus, Elias D.~Striatum},
pdfkeywords={First keyword, Second keyword, More},
}

\begin{document}
\maketitle

\begin{abstract}
	Epidemic modeling, encompassing deterministic and stochastic approaches, is vital for understanding infectious diseases and informing public health strategies. This research adopts a prescriptive approach, focusing on reinforcement learning (RL) to develop strategies that balance minimizing infections with maximizing in-person interactions in educational settings. We introduce SafeCampus \footnote{\url{https://github.com/ANRGUSC/SafeCampus}}, a novel tool that simulates infection spread and facilitates the exploration of various RL algorithms in response to epidemic challenges. SafeCampus incorporates a custom RL environment, informed by stochastic epidemic models, to realistically represent university campus dynamics during epidemics. We evaluate Q-learning for a discretized state space which resulted in a policy matrix that not only guides occupancy decisions under varying epidemic conditions but also illustrates the inherent trade-off in epidemic management. This trade-off is characterized by the dilemma between stricter measures, which may effectively reduce infections but impose less educational benefit (more in-person interactions), and more lenient policies, which could lead to higher infection rates.
\end{abstract}

\section{Introduction}
\label{sec:intro}
The progression back to everyday activities in this post-pandemic era emphasizes the critical importance of safety measures in educational spaces, a theme that resonates with the broader impacts of an epidemic. As the COVID-19 pandemic has shown \cite{world2020covid}, the intersection of epidemics and education demands innovative solutions to navigate these challenges, especially in balancing public health with the continuity of educational activities. This paper situates itself within this context, examining the role of reinforcement learning (RL) in enhancing safety measures on campuses during such public health crises. Reinforcement learning has been widely applied across various domains like healthcare, economics, and mobility for epidemic control. It employs algorithms like Deep Q-Learning, Proximal Policy Optimization, and others, alongside compartmental and more granular epidemic models (SEIR, SIR, SIHR, Meta-population, Agent-based) to simulate disease spread and optimize policies (\cite{arango2020covid}, \cite{ohi2020exploring}, \cite{feng2022precise}).
Building on existing research, this research integrates insights from works that have explored the impact of various interventions in educational settings during epidemics. For instance, \cite{oikawa2022class} provides valuable findings on the effectiveness of reducing class sizes to enhance social distancing and its consequential reduction in flu outbreaks. This aligns with our exploration of safety measures in schools, although our approach pivots towards the use of RL for broader epidemic management. Similarly, the studies by \cite{fukumoto2021shut} and \cite{wu2022global} delve into the effects of school closures on COVID-19 spread and the implications of such measures on children and society. Their work emphasizes the complexity of policy decisions in educational contexts during pandemics, a theme that resonates with our aim to identify strategies using reinforcement learning.

Furthermore,\cite{kaiser2020social} examines the efficacy of cohorting strategies within schools, emphasizing the need for well-thought-out interventions to minimize virus transmission, an aspect that our paper seeks to address through simulation-based policy evaluation. \cite{endo2022simulating} also contribute to this discourse by evaluating the efficacy of various school-based interventions, highlighting the potential limitations of certain approaches. 

Our contribution lies in creating SafeCampus, a tool that integrates various RL algorithms with stochastic classroom-based epidemic models to simulate disease spread on university campuses. This tool focuses on learning policies based on student attendance percentages to balance campus operations with health safety. Additionally, we identify policies that highlight the delicate balance between institutional operations and health safety

\section{Modeling}

\subsection{Problem definition}
Consider a classroom scenario with $N$ students attending sessions over $W$ weeks. During an ongoing epidemic, students face the risk of infection both off-campus and on-campus. Off-campus infections are considered to be an exogenous random process, where each student has an independent and identical probability  $c_w$ of being infected off-campus during week $w$, termed the community risk. On-campus infections result from infected students spreading a virus to other students. 

\subsection{Model Formulation}
We detail the development of an approximate SI (Susceptible, Infected) model designed to estimate the potential number of student infections under varying scenarios. It incorporates three primary inputs: the current number of infected individuals, the policy decision regarding the permissible number of people allowed in a given space, and the community risk $c_w$. This model is particularly useful in situations where the focus is on understanding the general spread of infection over time within a larger population. It could be ideal where quick, straightforward estimations of infection spread within individual classroom settings are needed.

\subsubsection{Approximate SI (Susceptible-Infectious) Model}
SI models involve differential equations to describe the rates at which individuals move between being susceptible and Infectious. This approximation instead uses a simpler algebraic approach for more straightforward to implement and interpret in the context of a classroom. We formulate an approximate SI model to estimate the number of students infected as follows:

Let $N_i$ represent the number of students allowed per course for the $i$-th course, and $I_i^{(c)}$ denote the current number of infected individuals in that course. The community risk of infection is represented by $c_{\text{risk}}$. The number of new infections $I_i^{(n)}$ in each course is then estimated using the following relation:

\begin{equation} \label{eq:approximate_model}
I_i^{(n)} = \min\left(\left[\alpha_m \cdot I_i^{(c)} \cdot N_i + \beta \cdot 
 c_{\text{r}} \cdot N_i^2\right], N_i\right)
\end{equation}

where:
\begin{itemize}
    \item $\alpha_m$ reflects the transmission risk associated with an infected individual within the classroom and $\beta$ scales the effect of the community risk ( $c_r$) on the infection rate, taking into account the probability of students getting 
    infected outside and contributing to the spread within the classroom.

\end{itemize}
Now let $\mathcal{I}_w$ represent the set of students infected in week $w$, either off-campus or on-campus. We assume that this is the set of students that are potential spreaders for week $w+1$. Rather than tracking individual students, we focus on the aggregate number of infected students  $I_w = |\mathcal{I}_w|$. $I_w$. Also the output of the expected models that are to interact with the SafeCampus environment, $I_w$ is inherently a random variable, but in this simple model we track $E[I_w]$, the expected value of $I_w$.

\section{The Reinforcement Learning Problem}

We formulate the problem of finding operational strategies during an ongoing epidemic as a model-free RL problem where we are interested in developing a policy that would take a given set of observations about the infection process and make a decision on how many students to allow in the classroom at the beginning of every week $w$. The RL problem is thus formulated as follows:
\begin{itemize}
    \item \textbf{State space: }We use as the state observation a tuple consisting of the community risk and the current number of (expected) infected students, i.e. ($c_w$, $E[I_w]$). For simplicity and efficiency, we are currently discretizing the observed state space into a set of discrete levels for both $c_w$, $E[I_w]$. A range of 1-10 is used and this could be easily modified to accommodate a more fine-grained discretization at the expense of greater storage and computational complexity for the reinforcement learning.
    
    \item \textbf{Action space: }The output of the policy $A_w$ is the number of students allowed to participate in the class. Again, for ease of implementation, we discretize the action into $L$ levels (e.g., if $L=3$, the possible actions may be to allow $0$, $0.5N$ or $N$ students in a given week).
    \item \textbf{State-transition model: }  Within our environment, this is governed by the dynamics of the approximate SI model described earlier which simulates the spread of infection in a classroom setting. 
    \item \textbf{Reward: } This is designed to consider community risk, the number of allowed students, and the current number of infected students. It is defined as:
        \begin{equation}
            \text{reward} = \text{int}\left( \alpha_r \times \text{allowed\_students} - ((1 - \alpha_r) \times \text{current\_infected\_students}) \right)
        \end{equation}

        where:
        \begin{itemize}
            \item $\text{current\_infected\_students}$ is the sum of infected students.
            \item $\text{allowed\_students}$ is the sum of students allowed per course.
            \item $\alpha_r$ is a weighting factor.
        \end{itemize}
\end{itemize}

\section{Software Architecture Overview}
\begin{figure}[h] 
\includegraphics[width=\columnwidth]{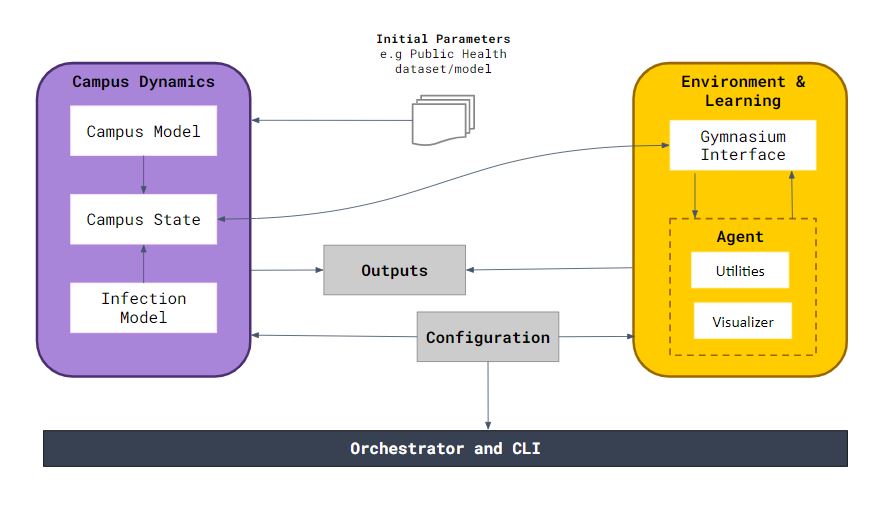}
\caption{A schematic of the system, integrating campus dynamics, agent learning mechanisms, and orchestration components.}
\label{fig:architecture}
\end{figure}

 SafeCampus is engineered to simulate a range of scenarios, enabling the evaluation of various policy decisions concerning infection control and in-person interactions. Figure \ref{fig:architecture} shows the Key components that includes the Campus Dynamics modules, which are essential in defining and managing the evolving state of the campus environment. This includes the \texttt{campus model}, \texttt{campus state}, and \texttt{infection model}, each responsible for initializing the simulation, managing dynamic system states, and simulating infection spread using various epidemiological models. The Environment and Learning component integrates these behaviors to optimize strategies through learned experiences in an RL framework, utilizing tools like the \texttt{gymnasium interface} for agent interactions and an \texttt{agent package} for implementing various RL algorithms, primarily focusing on Q-learning. Supplementary modules, such as the \texttt{configuration}, \texttt{outputs}, and \texttt{orchestration} modules, provide essential support in system configuration, data management, and overall system control, respectively. Furthermore, the \texttt{command-line interface (CLI)} facilitates user interaction, allowing for precise control over operational modes and parameters, thus driving the system's training, evaluation, and optimization processes. We discuss the implementation in Appendix \ref{appendix:software}

\section{Experiments}
In our study, we hypothesize that Q-learning can effectively generate a sensible policy matrix that prescribes specific occupancy decisions based on infection counts and community risk patterns. We also posit that the reward weight parameter \(\alpha_r\) will play a crucial role in shaping the Q-learning agent outcomes and the precision of the policy matrix. By varying the \(\alpha_r\) value for training, our objective is to explore how the algorithm prioritizes educational benefits by 'allowing more students' and infection risk minimization, thereby calibrating the matrix to align with varying epidemic scenarios. 

\section{Results}
\begin{figure}
    \centering
    \includegraphics[width=0.8\textwidth]{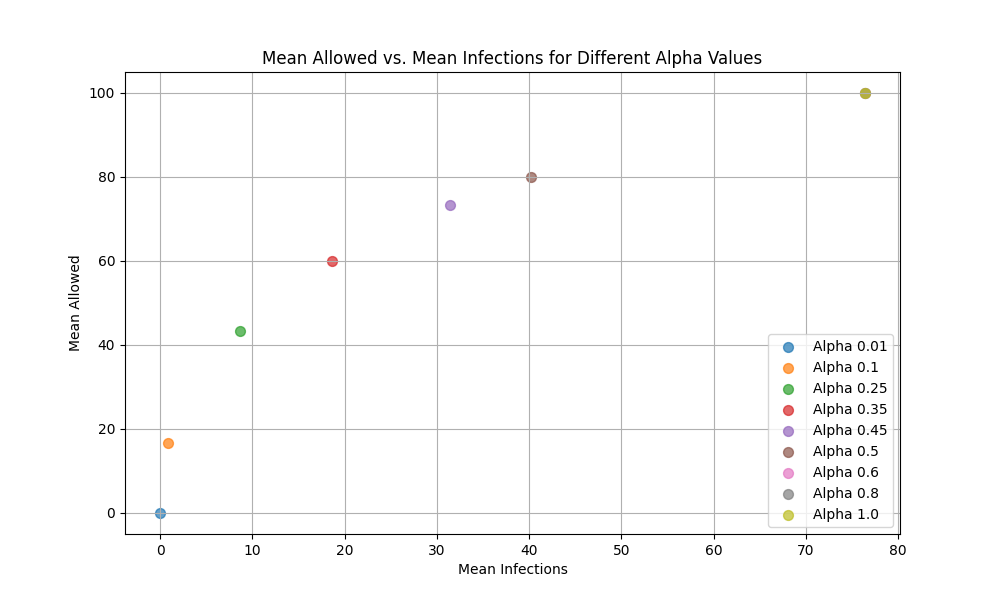}
    \caption{Trade-off between occupancy levels and infection rates across various alpha-driven policies. Each point represents the outcome of a policy, with color coding indicating different alpha values }
    \label{fig:tradeoff}
\end{figure}

\begin{figure}
    \centering
    \subfigure[\(\alpha_r\) = 0.01]{\includegraphics[width=0.28\textwidth]{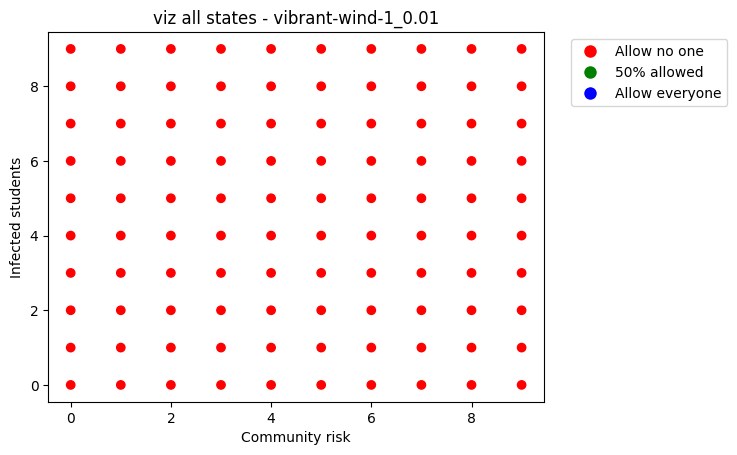}}
    \subfigure[\(\alpha_r\) = 0.1]{\includegraphics[width=0.28\textwidth]{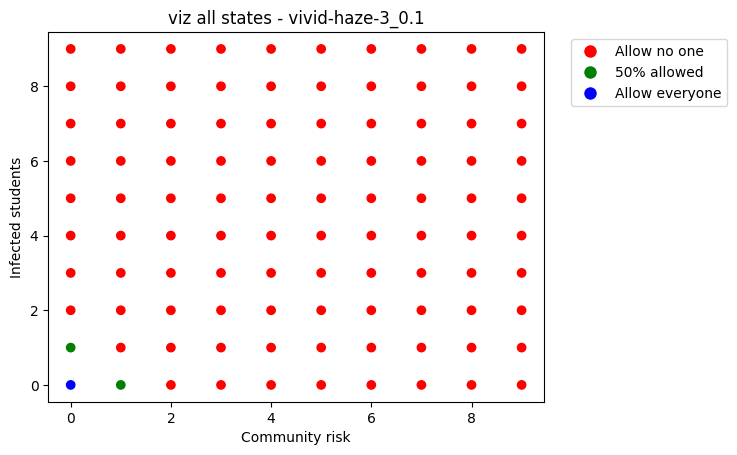}} 
    \subfigure[\(\alpha_r\) = 0.25]{\includegraphics[width=0.28\textwidth]{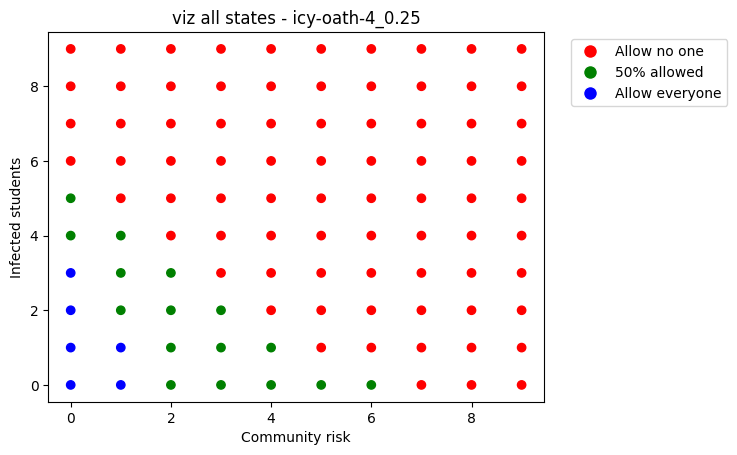}} 

    \subfigure[\(\alpha_r\) = 0.35]{\includegraphics[width=0.28\textwidth]{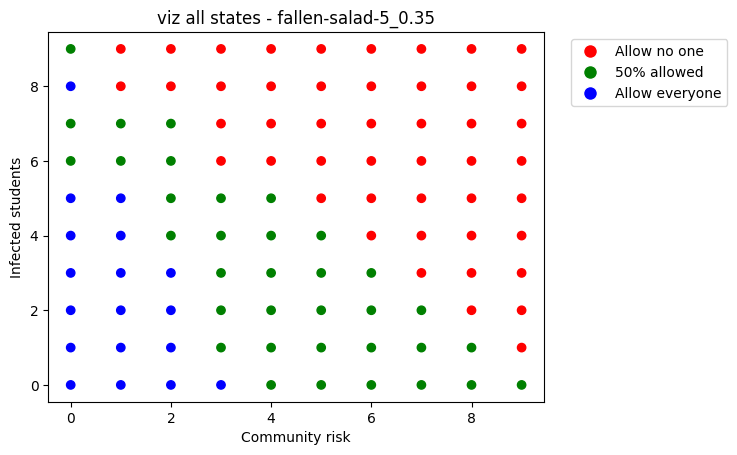}}
    \subfigure[\(\alpha_r\) = 0.45]{\includegraphics[width=0.28\textwidth]{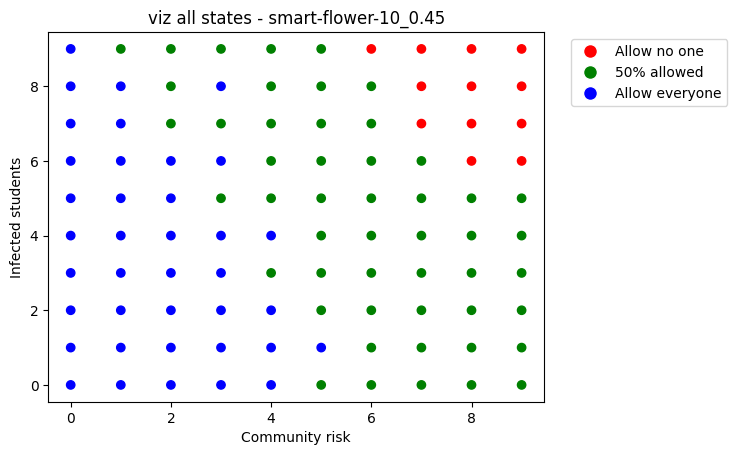}} 
    \subfigure[\(\alpha_r\) = 0.5]{\includegraphics[width=0.28\textwidth]{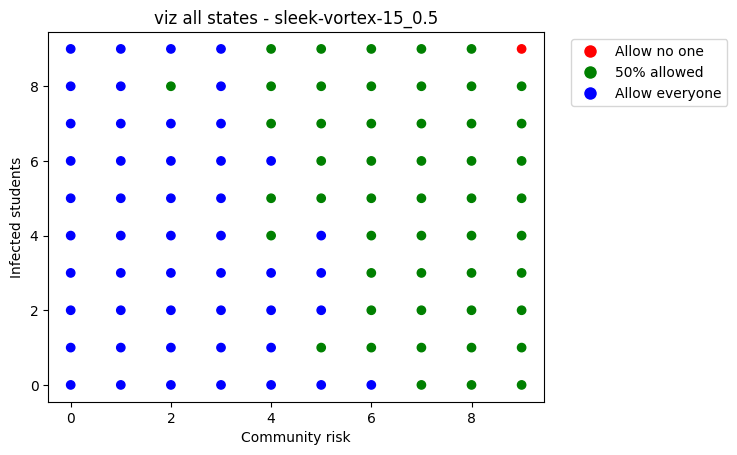}} 
    
    \subfigure[\(\alpha_r\) = 0.6]{\includegraphics[width=0.28\textwidth]{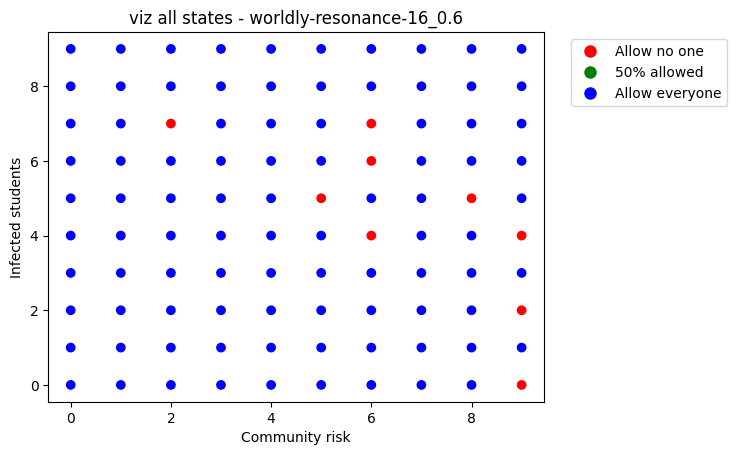}}
    \subfigure[\(\alpha_r\) = 0.8]{\includegraphics[width=0.28\textwidth]{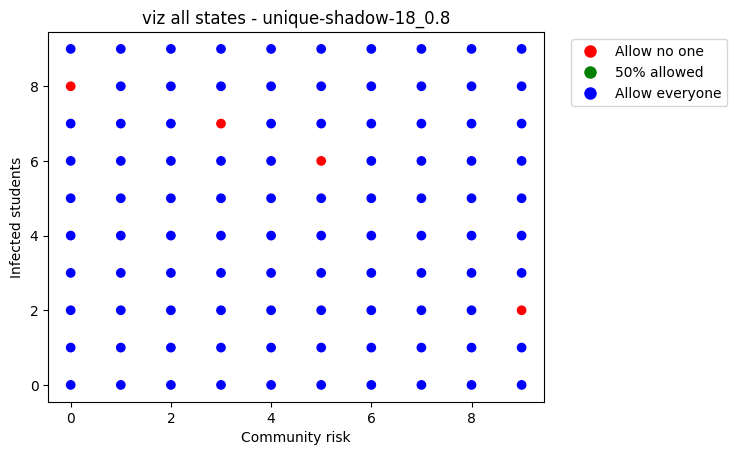}} 
    \subfigure[\(\alpha_r\) = 1.0]{\includegraphics[width=0.28\textwidth]{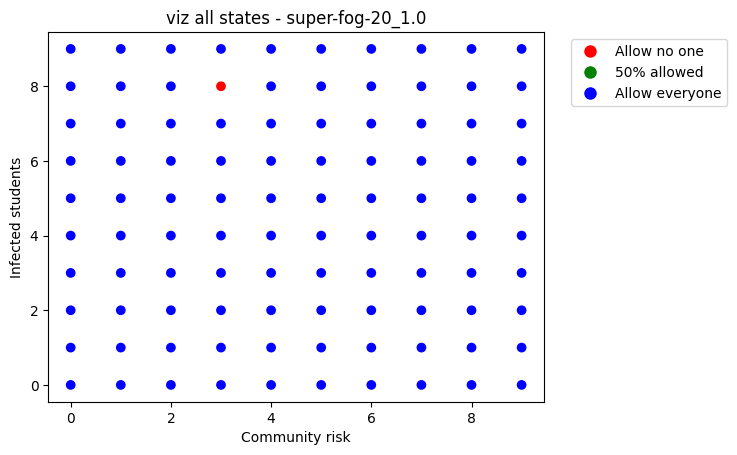}}
    \caption[Policy Matrices with Varying \(\alpha_r\)]{(a) - (i) illustrate the policy matrices generated by the Q-learning algorithm under varying values.}
     \label{fig:policies_matrices}
\end{figure}

In Figure \ref{fig:policies_matrices}, each matrix depicts the algorithm's occupancy recommendations across different states of community risk and infection counts. The policy gradient shifts from conservative (red dots: Allow no one) to permissive (blue dots: Allow everyone) as the alpha value increases, indicating a higher emphasis on educational benefits. Green dots represent a balanced occupancy decision (50\% allowed). The progression from (a) to (i) captures the algorithm's adaptive responses to the campus dynamics, showcasing the delicate balance between ensuring educational benefits and managing infection risks. Figure \ref{fig:tradeoff} shows the scatter plot for evaluation data for the policies learned during the training phase for different \(\alpha_r\) values. Lower \(\alpha_r\) values tend to yield points clustered towards the left of the plot, indicating a propensity for policies that favor minimizing infection risk, often resulting in reduced occupancy. Conversely, higher \(\alpha_r\) values shift the balance towards maximizing occupancy and educational benefits, leading to points situated towards the right, which accept higher infection rates. These data points reflect the agent's adaptive policy decisions under different reward prioritization.

\section{Conclusion and Future Work}
We demonstrate the application of reinforcement learning, specifically Q-learning in developing adaptable policies for managing operational decisions during an epidemic. In our case,  striking a balance between educational benefits and health risks. Furthermore, the implementation of our software could serve as a collaborative platform bridging the gap between epidemiologists,  artificial intelligence researchers, policymakers, and administrators, fostering an integrative environment for data-driven decision-making and policy formulation in response to public health challenges. The policy outcomes for each \(\alpha_r\) value also reveal the variability of the agent's decisions, influenced by the stochastic nature of the simulated environment. This evaluative approach not only enhances our understanding of the Q-learning agent's capabilities but also emphasizes the importance of collaborative decision-making with domain experts to navigate complex trade-offs effectively. For the future, we would like to explore other types of stochastic epidemic models especially those tailored for indoor environments such as \cite{hekmati2022simulating} which is designed to capture the dynamics of disease transmission in settings like classrooms as well as examining other model-free algorithms including deep reinforcement learning methods to assess the effectiveness of these algorithms in our tool.

\appendix
\clearpage
\begin{table}[hbtp]
\centering
\footnotesize
  {tab:rlepidemic-control}
  {\caption{Related works on Reinforcement Learning applied to Epidemic Control}}
  {\begin{tabular}{p{0.25\linewidth}p{0.35\linewidth}p{0.35\linewidth}}
  \toprule
  \bfseries{RL Approach} & \bfseries{Citation} & \bfseries{Optimization Goal} \\
  \midrule
  Deep (Q) Reinforcement Learning & 
\begin{tabular}{@{}l@{}}
    \cite{arango2020covid}\\
    ICU bed usage, \\
    \cite{bushaj2023simulation}\\
    Epidemic interventions, \\
    \cite{deng2021optimal}\\
    Infection rates, economic impact, \\
    \cite{guo2022pacar}\\
    Infection spread, economic impact, \\
    \cite{khadilkar2020optimising}\\
    Disease spread, economic impact, \\
    \cite{khatami2022deep}\\
    Epidemic, economic impact, \\
    \cite{ohi2020exploring}\\
    Economic ratio, death ratio, \\
    \cite{probert2019context}\\
    Livestock disease outbreak, \\
    \cite{wang2023platform}\\
    Economic platform behavior
\end{tabular} & 
\begin{tabular}{@{}l@{}}
    Minimize ICU bed usage overshoots, \\
    Compare vaccination strategies, \\
    Maximize economic output, control infections, \\
    Minimize infection spread and restrictions, \\
    Balance health and economic costs, \\
    Minimize epidemic and economic burdens, \\
    Balance pandemic and economic situation, \\
    Develop state-dependent response policies, \\
    Promote efficiency and resilience\\
\end{tabular} \\
\\[-0.9ex]
Proximal Policy Optimization & 
\begin{tabular}{@{}l@{}}
    \cite{feng2023contact}\\
    Infections, mobility cost, \\
    \cite{feng2022precise}\\
    Spread of epidemic, \\
    \cite{hosseinloo2022data}\\
    Indoor viral exposure, \\
    \cite{libin2021deep}\\
    Pandemic influenza spread\\
\end{tabular} & 
\begin{tabular}{@{}l@{}}
    Minimize infections and mobility cost, \\
    Control epidemic spread, optimize interventions, \\
    Minimize occupants' exposure to pathogens, \\
    Learn prevention strategies
\end{tabular} \\
 \\[-0.9ex]
Hierarchical Reinforcement Learning &
\begin{tabular}{@{}l@{}}
    \cite{feng2022precise}\\
Spread of epidemic
\end{tabular} & 
\begin{tabular}{@{}l@{}}
   Control epidemic spread\\
   optimize mobility interventions \\
\end{tabular} \\

\\[-0.9ex]
Reinforcement Learning & 
\begin{tabular}{@{}l@{}}
    \cite{kompella2020reinforcement}\\
Hospital capacity, economic impact 
\end{tabular} & 
\begin{tabular}{@{}l@{}}
   Optimize mitigation policies\\
   to reduce economic impact \\
\end{tabular} \\

 \\[-0.9ex]
Various RL Algorithms & 
\begin{tabular}{@{}l@{}}
   \cite{uddin2020optimal}\\
Testing, sanitization,lockdown levels\\
Hospital capacity, economic impact 
\end{tabular} & 
\begin{tabular}{@{}l@{}}
   Control virus spread, optimize resource usage \\
\end{tabular} \\
 \\[-0.9ex]
Gaussian Process regression, Stochastic multi-armed bandits & 
\begin{tabular}{@{}l@{}}
   \cite{bent2018novel}\\
Malaria policy 
\end{tabular} & 
\begin{tabular}{@{}l@{}}
   Find optimal malaria policy \\
\end{tabular} \\

  \bottomrule
  \end{tabular}}
\end{table}
\section{More on Approximate SI Model}
\subsection{Assumptions}
The assumptions below facilitate a more straightforward and computationally less intensive approach, however, they also delineate the limitations of the model. The validity of the model's predictions is contingent on the extent to which these assumptions hold true.
\begin{itemize}
    \item \textbf{Simple Infection Dynamics}: The term \(\alpha \cdot I_i^{(c)} \cdot N_i\), simplifies the complex dynamics of disease transmission, omitting potentially more complex non-linear interactions and feedback mechanisms that could be present in a real-world setting.
   
    \item \textbf{Constant Transmission and Community Risk Parameters}: The constants \(\alpha\) and \(\beta\) are assumed to be fixed values, representing the transmission risk associated with an infected individual in the classroom and the effect of community risk on infection rates, respectively. This assumption does not account for temporal or situational variations in these parameters, such as changes in community transmission rates or the implementation of mitigation measures.
    \item \textbf{Upper Bound on Infections}: The use of the \(\min(\cdot)\) function in the model imposes an upper limit on the predicted number of new infections, ensuring it does not exceed the total number of students in a course. 
    \item \textbf{Exclusion of Recovery and Immunity Dynamics}: Unlike the compartmental models such as the SIR, this approximation does not explicitly model the transition of individuals to 'Recovered' status or consider the effects of immunity, either natural or vaccine-induced. This omission simplifies the model but reduces its ability to capture the full spectrum of disease progression and its impact over time.
    \item \textbf{Isolated Classroom Consideration}: The model treats each classroom as an isolated unit, without considering cross-class interactions or the broader campus-wide network of social contacts. This assumption overlooks the potential for inter-class transmission and the cumulative effect of multiple infection sources.
\end{itemize}

\subsection{Model Dynamics}
We design a function in Python to execute a permutation of scenarios by iterating over predefined lists of initial infected cases, allowed class sizes, and community risk. In each iteration, it calculates the number of infected students using the model. The function dynamically updates the current infected count for each iteration, using the previous iteration's output as the next input. The output is a collection of data points, each representing a unique combination of these three parameters—current infected count, class size, and community risk—and the corresponding calculated number of infected students. 

\begin{figure}[ht]
    \centering
    \begin{minipage}{0.5\textwidth}
        \centering
        \includegraphics[width=\textwidth]{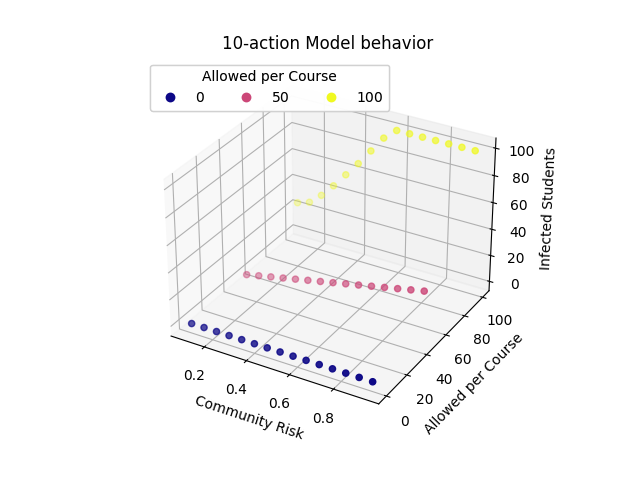} 
    \end{minipage}\hfill
    \begin{minipage}{0.5\textwidth}
        \centering
        \includegraphics[width=\textwidth]{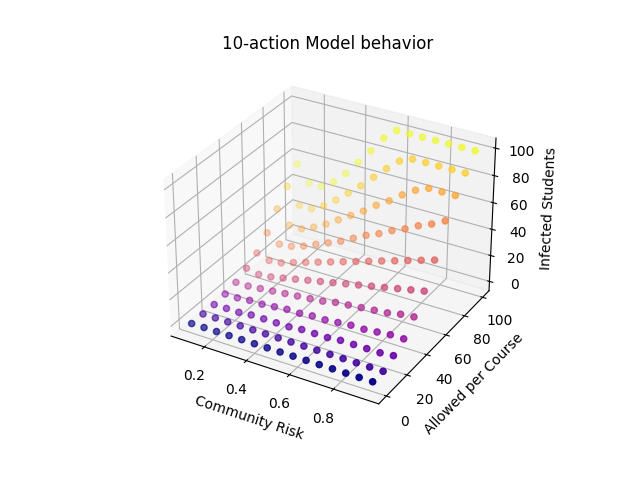} 
    \end{minipage}
     
   \caption {Simulation results from the function, depicting infection spread in a classroom setting. On the left, data from simulations with allowed class sizes of 0, 50, and 100, and on the right, a more granular range from 0 to 100 in increments of 10. Each data point represents a combination of initial infected cases, occupancy level, and community risk.}
   \label{fig:simulation}
  
\end{figure}

Figure \ref{fig:simulation} provides a view of the model’s behavior. There is a correlation between increasing class size and the number of current infections with a subsequent rise in predicted infections. This trend is consistent across all occupancy levels, highlighting the model's sensitivity to both initial infection levels and population density in a classroom setting. The non-linear escalation in the number of predicted infections with larger occupancy levels is particularly pronounced, suggesting an exponential relationship between occupancy levels and infection risk within the model’s framework. Additionally, the influence of varying community risk levels on the number of predicted infections is evident across all occupancy levels. Higher community risk levels correlate with increased infection predictions, underlining the model's incorporation of external epidemiological factors. The saturation effect observed in both sets of data, where the number of predicted infections plateaus at higher infection levels, especially in larger class sizes, indicates an upper limit imposed by the model. 

\section{Q-Learning}

We apply Q-learning in this context since it does not require a predefined model of the environment's dynamics, making it suitable for situations where the exact mechanisms of infection spread are complex or not fully understood. Q-learning learns from interactions with the environment, gaining knowledge directly through trial and error. This allows for adaptability and flexibility in evolving scenarios like infection spread, where variables such as individual behavior, policy changes, and external factors can alter the dynamics unpredictably. The learning process involves updating the Q-values using the observed rewards and the estimated future values. The algorithm progressively shifts from exploration to exploitation, initially exploring the action space widely and gradually honing in on the best strategies. The use of a high initial Q-value (optimistic initialization) encourages initial exploration, prompting the agent to try out various actions to discover their potential. Over time, as the agent learns the consequences of different actions, it starts to exploit this knowledge to make more informed decisions. The balance between exploration and exploitation is maintained through the decay of exploration and learning rates, allowing the agent to adapt its strategy based on accumulated knowledge and experience. \\
\clearpage
\begin{algorithm}
\caption{Optimistic Q-Learning}
\label{alg:q_learning}
\KwIn{Action space, state space, hyperparameters}
\KwResult{Optimized policy for decision making}
Initialize Q-table with high values\;
\For{episode $\leftarrow 1$ \KwTo max\_episodes}{
    state $\leftarrow$ reset environment\;
    terminated $\leftarrow$ False\;
    \While{not terminated}{
        Choose an action based on current policy\;
        Execute action and observe reward, next state\;
        Update Q-table according to:\\
        \tcp{Q-learning update equation:}
        \tcp{$Q(s, a) \leftarrow Q(s, a) + \alpha \left[r + \gamma \max_{a'} Q(s', a') - Q(s, a)\right]$}
        Update state to next state\;
    }
    Update policy based on learned Q-values\;
    \tcp*[r]{Decay exploration and learning rates as needed}
}
\end{algorithm}
\section{Training}
\subsection{Experiment setup}
The training and evaluation were conducted on a system running Ubuntu 22.04.3 LTS with an x86\_64 architecture, powered by an 11th Gen Intel(R) Core(TM) i7-11800H processor clocked at 2.30GHz. Python environment management was handled using Conda, with Python version 3.8 or higher. For tracking and logging experiment details, Weights \& Biases (wandb)\footnote{See \url{https://wandb.ai/site} for more information on wandb.} was integrated into the setup.

\subsection{Hyperparameters}
The following hyperparameters were utilized:
\begin{table}[h!]
\centering
\begin{tabular}{|l|l|}
\hline
\textbf{Parameter} & \textbf{Value} \\ \hline
Learning Rate & 0.3 \\ \hline
Discount Factor & 0.9 \\ \hline
Max Episodes & 2500 \\ \hline
Exploration Rate & 1.0 \\ \hline
Min Exploration Rate & 0.000001 \\ \hline
Exploration Decay Rate & 0.001 \\ \hline
learning\_rate\_decay & 0.99999 \\ \hline
min\_learning\_rate & 0.001 \\ \hline
\end{tabular}

\caption{Hyperparameters Configuration}
\end{table}
\subsection{Performance}
\begin{figure}
    \centering
    \includegraphics[width=0.8\textwidth]{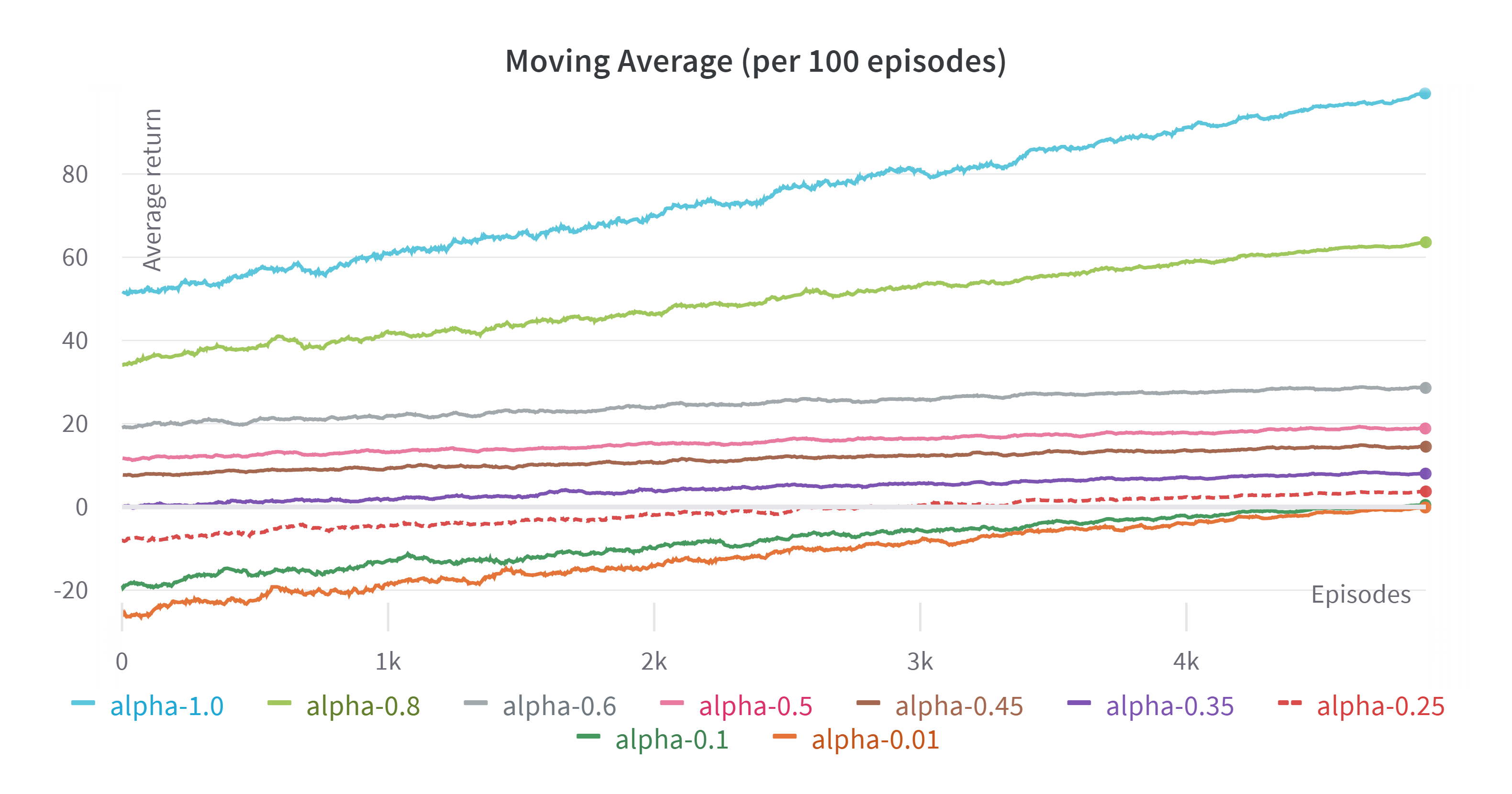}
    \caption{Moving Average showcasing the evolution of the Q-learning algorithm's performance with varying the reward weight \(\alpha\) values over 2500 training episodes. }
    \label{fig:moving-average}
\end{figure}

Figure \ref{fig:moving-average} shows the curves representing the moving averages of the expected return(reward) with lower \(\alpha_r\) values corresponding to the lower part of the graph and higher values with the upper part. This distinction suggests that the choice of \(\alpha\) significantly impacts Q-learning performance. 

\subsection{State Visitation}
\begin{figure}
    \centering
    \subfigure[\(\alpha_r\) = 0.01]{\includegraphics[width=0.31\textwidth]{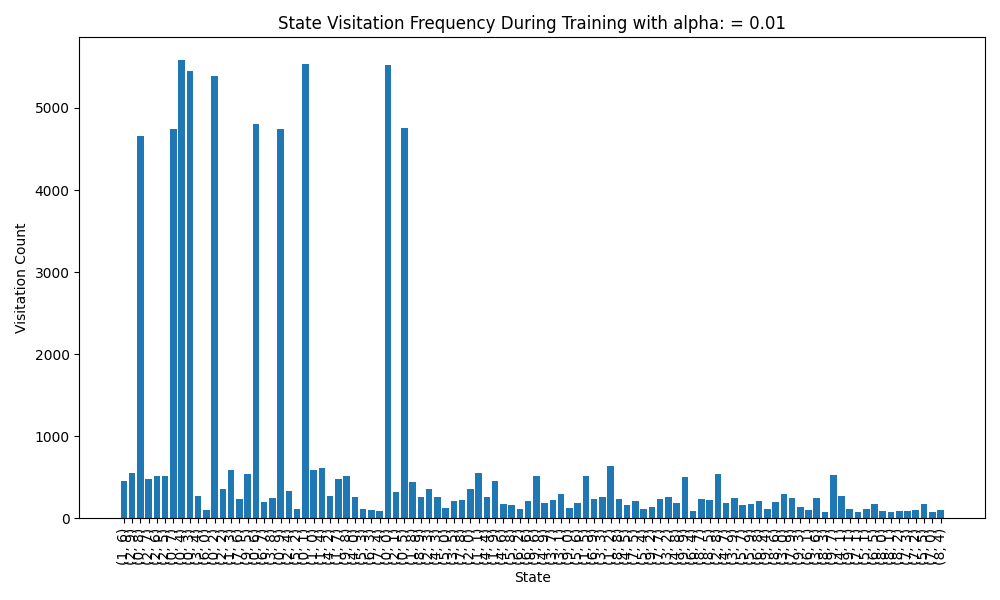}}
    \subfigure[\(\alpha_r\) = 0.1]{\includegraphics[width=0.31\textwidth]{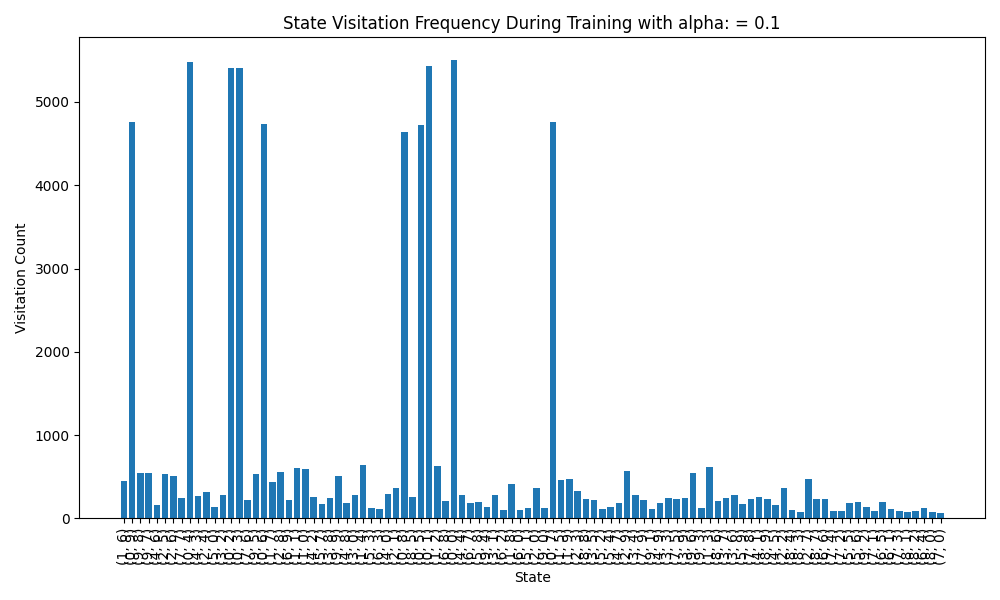}}
    \subfigure[\(\alpha_r\) = 0.25]{\includegraphics[width=0.31\textwidth]{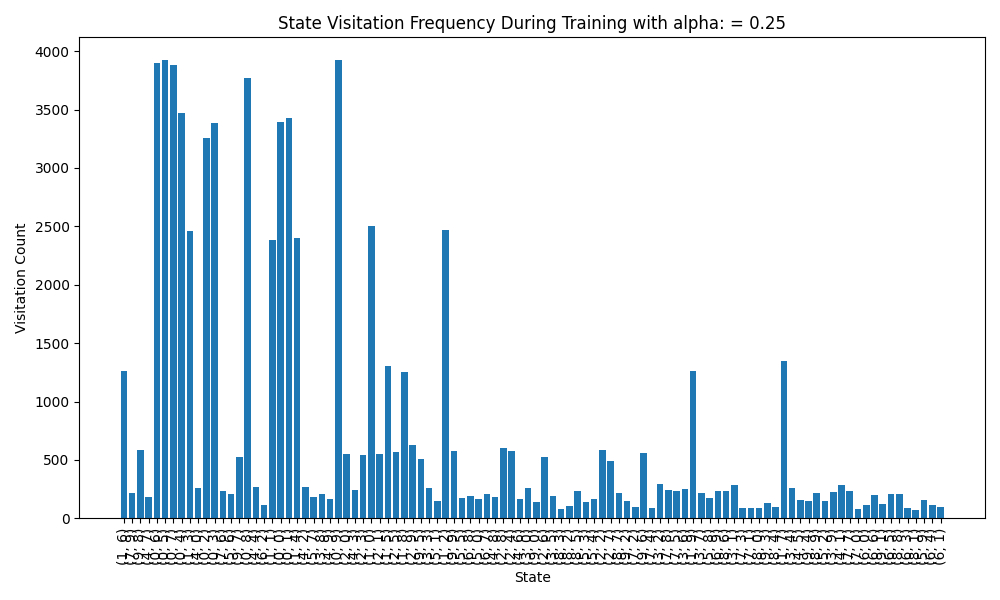}}

    \subfigure[\(\alpha_r\) = 0.35]{\includegraphics[width=0.31\textwidth]{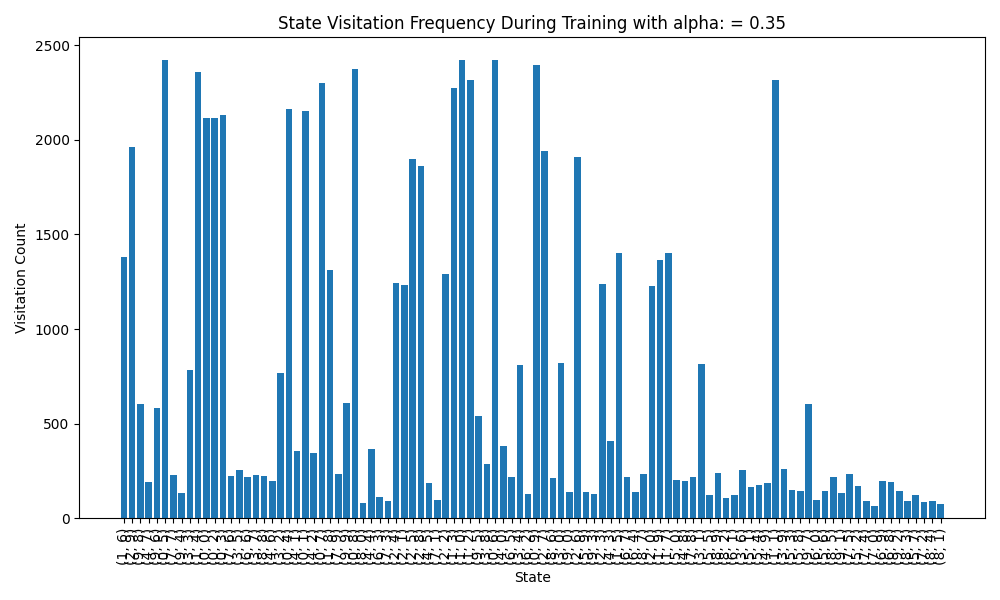}}
    \subfigure[\(\alpha_r\) = 0.45]{\includegraphics[width=0.31\textwidth]{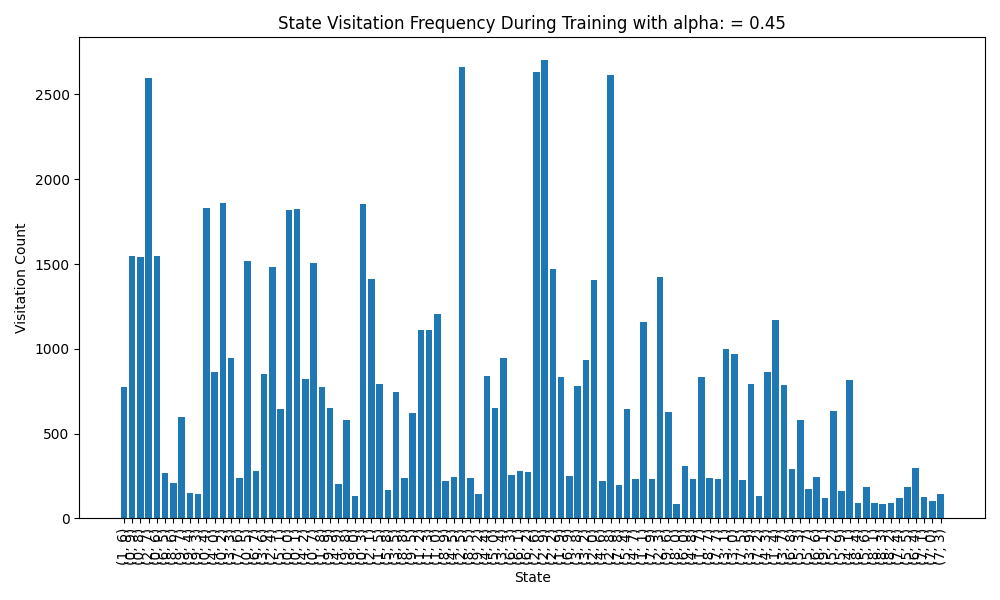}}
    \subfigure[\(\alpha_r\) = 0.5]{\includegraphics[width=0.31\textwidth]{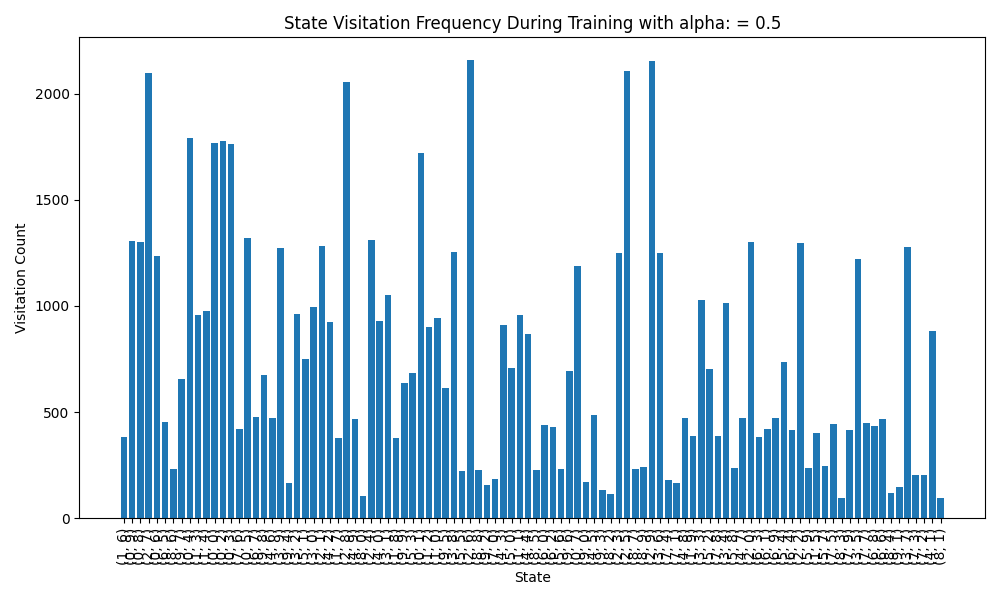}}

    \subfigure[\(\alpha_r\) = 0.6]{\includegraphics[width=0.31\textwidth]{figures/alpha-5.png}}
    \subfigure[\(\alpha_r\) = 0.8]{\includegraphics[width=0.31\textwidth]{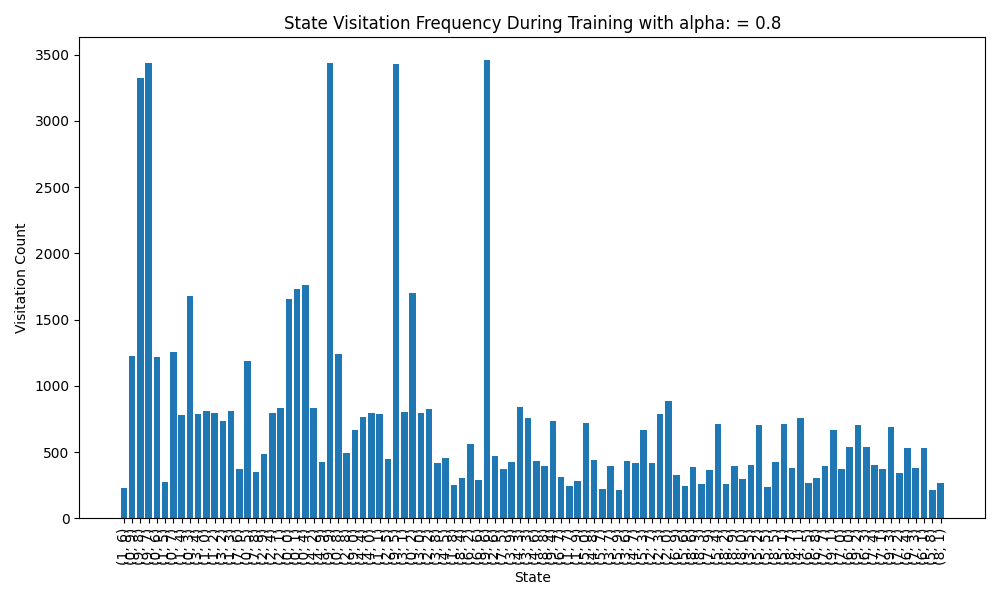}} 
    \subfigure[\(\alpha_r\) = 1.0]{\includegraphics[width=0.31\textwidth]{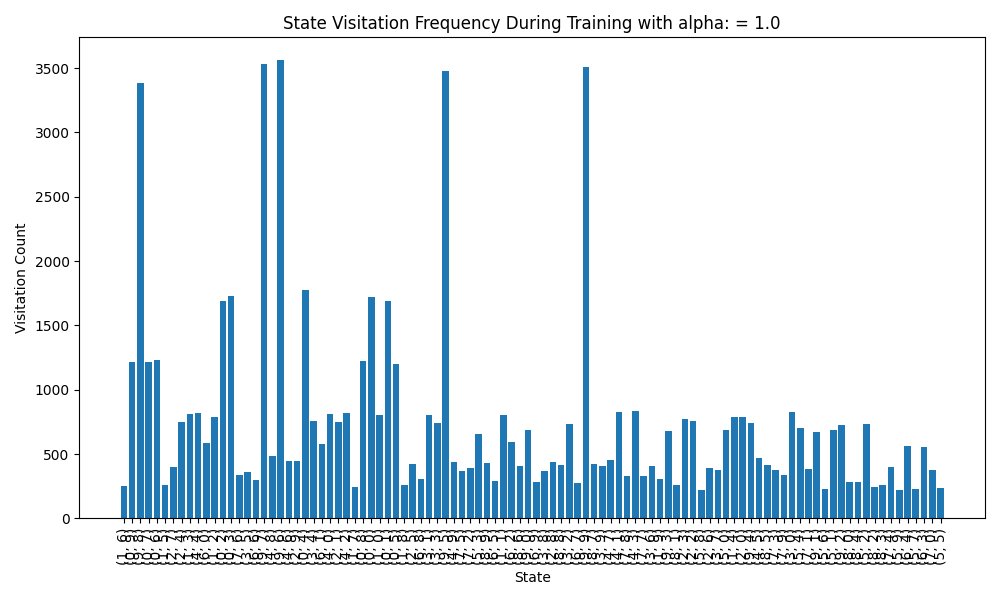}}
    
    \caption[]{State Visitations for the different Alpha Values during training}
     \label{fig:state-visits}
\end{figure}

The state visitation frequency histograms as shown in Figure \ref{fig:state-visits} provide insight into how often each state was visited during training. A well-distributed visitation frequency across states suggests a thorough exploration of the state space, which is crucial for the algorithm to learn a near-optimal policy. However, if the visitation frequency shows that certain states are visited disproportionately, it indicates that the algorithm may have biases in its exploration or that some states are inherently more common given the dynamics of the simulated environment. Ideally, we want a balanced state visitation where all states are sufficiently explored. This ensures that the Q-learning algorithm has a comprehensive understanding of the environment and can make informed decisions for each state. If the visitation frequency is skewed, with some states rarely visited, the algorithm may not learn how to act optimally in those states, potentially leading to overfitting on more common states. This can result in a policy that is not generalizable to all possible scenarios. We use this as a  diagnostic tool in understanding the learning process of the Q-learning algorithm. It helps us identify whether the exploration strategy is effective and whether the algorithm is learning to focus on the most rewarding states, both of which are necessary for developing a refined and almost perfect policy matrix.
\clearpage
\section{More on Software Architecture}
\label{appendix:software}
\textbf{Initial parameters:} These are a set of parameters used to initialize the \textit{campus model}. They include the number of courses on the campus, the number of students enrolled per course, the duration or number of weeks the simulation will run, and the initial rate at which the students are infected. The campus model uses these initial parameters to set up the simulation environment. Specifically, the \textit{initial infection rate} can be dynamic, potentially sourced from external public health datasets, or updated based on real campus data to ensure the simulation reflects real-world conditions.

The \textbf{Campus Dynamics:} This component provides the structure and logic governing the campus environment and its interaction over time. It has the following modules:
\begin{itemize}
    \item \textbf{Campus Model} represents the campus environment in the simulation. It is initialized with parameters that define the layout and initial state of the campus, such as the number of courses, the number of students per course, the duration of the simulation in weeks, and the initial rate of infection. These parameters can be set deterministically or sourced from external data. The class provides methods to retrieve the deterministically generated number of students for each course, the total duration of the simulation, and the initial number of infected students per course based on the infection rate.
    \item \textbf{Campus State} functions as the dynamic state management module in the system's architecture Also functions as a state space wherein the system evolves over discrete time steps, governed by deterministic and stochastic transition functions.It interfaces directly with the campus model and it dynamically regulates, updates, and reflects the evolving state of the system based on a combination of predetermined actions and internal models. It utilizes state variables such as \textit{student\_status}, \textit{allowed\_ students\_per\_course}, and \textit{community\_risk} to represent the current state of the campus environment. The \textit{update\_with\_action (action)]} method integrates the input action into the state, applying state transition logic defined by methods like \textit{apply\_action()}. This transition logic is influenced by both the deterministic properties provided by the campus model and the stochastic processes associated with infection dynamics. The \textit{get\_reward (alpha)} method calculates a reward metric based on current state variables. This feedback mechanism is crucial for the reinforcement learning process to optimize actions. Lifecycle management is addressed through the \textit{is\_episode\_done ()} and \textit{reset()} methods which ensure that the system can initialize, iterate and conclude the reinforcement learning episodes, maintaining a consistent state representation throughout.
    \item The \textbf{Infection Model} encapsulates the logic to estimate the spread of infections in the campus environment. The \textit{ get\_infected \_students\_si} provides an estimate of the infected students using the approximate SI model
\end{itemize}
The \textbf{Environment and Learning} components for the system's decision-making mechanism, integrate the dynamic behaviors modeled in previous modules and optimize strategies based on learned experiences. 
\begin{itemize}
    \item The \textbf{Gymnasium Interface} implements the \textit{CampusGymEnv} class which is structured to leverage the capability of the Gymnasium framework (a maintained fork of OpenAI's Gym library). This is an API standard for reinforcement learning that makes it amenable for learning agents to interact with the dynamic system encapsulated by the campus model and its state. The environment uses a multi-discrete space for both actions and observations. This implies that each component of the space can take on one of several discrete values. For the CampusGymEnv, the action space is composed of multiple discrete slots, each corresponding to a course, where each slot can assume a value indicative of the percentage of students permitted to attend physically (e.g 0\%, 50\%, 100\%). The observation space on the other hand, comprises discrete values representing the number of infected students in each course and an additional value for community risk. This multi-discrete representation simplifies the complexity of the environment, enabling agents to operate on a finite set of states and actions, making the learning process more tractable. Dealing with a smaller state space is crucial in reinforcement learning, especially for methods like Q-learning that essentially tabulate values for each state-action pair. A reduced state space can significantly mitigate the curse of dimensionality, a challenge in which the state or action space grows exponentially with the addition of each new dimension. This exponential growth can make learning infeasible due to the sheer amount of data required.  consider a naive representation of our environment, where each student's health status (infected or not) is a separate dimension in our state space. Given $n$ students, the state space would be \(2^n\), which quickly becomes unmanageable even for a modest number of students. By discretizing or bucketing continuous or large discrete values into smaller categories, we effectively reduce the state space, making it more manageable for Q-learning or similar tabular approaches. This discretization strategy can be likened to a form of state aggregation, where multiple states are treated as a single state by the learning algorithm, thus reducing the complexity. We utilize helper functions to navigate between continuous or extensive discrete realms and their condensed counterparts. These functions transform raw values into a more compact state space, to make learning more tractable, efficient, and resource-conservative, a critical consideration for real-world, dynamic systems. Core methods like \textit{step(), reset() and render()} underpin the environment's operation. The step drives the system dynamics based on agent actions, reset reinitializes the environment, crucial for episodic RL tasks. The render method, although primarily diagnostic in this implementation, offers potential visual insights into the environment's state.
    \item The \textbf{Agent package} offers a blueprint for integrating a myriad of RL algorithms. The default choice is a Q-learning agent. The modular design choice allows for researchers and developers to incorporate alternative agents, tailoring the system's learning mechanism to specific requirements or experimental paradigms. Regardless of the chosen agent strategy, a few components remain invariant. The \textit{Utilities} module, for instance, offers essential functions that facilitate the ingestion of configuration parameters and other vital attributes for the agent. Simultaneously, the \textit{Visualizer}module provides a suite of visualization functions that cater to any agent's trajectory. This makes its easier to interpret the agent's actions and decision-making process. These visualizations include the granular view of the agent's evolving Q-values which represent its understanding of the values of actions in given states. The variance in rewards provides insights into the learning stability and convergence which is crucial for diagnosing training inefficiencies or anomalies. For the default q-learning agent, we have a  function \textit{visualize\_all\_states} which provides a visual representation of the agent's decision-making strategy or policy across the discretized states. For each state, defined by a combination of community risk and infected students, the function determines the optimal action by consulting the Q-table, a learned matrix that captures the expected rewards for each state-action pair. The resultant visualization offers a scatter plot where points represent different states and their colors indicate the chosen action. This allows for an immediate interpretation of the agent's behavior in diverse scenarios, illuminating its tendencies and potential areas of concern in its decision-making process.
\end{itemize}
The \textbf{Configuration} module is responsible for managing and loading configurations required for the entire system. These include specific agent hyperparameters like learning rates, environment constants, episode lengths, and shared configurations such as file paths for results. Externalizing these settings into a configuration module, allows for flexible experimentation, as researchers and developers can modify parameters without altering the core codebase.
\textbf{Outputs} module handles the collection, storage, and retrieval of results generated during the agent training and evaluation.

The \textbf{Orchestration} module is designed to control and manage the operations of the system. It primarily functions by interpreting configuration files and executing tasks based on these configurations. Within the orchestrator, there are distinct functions for different operations: \textit{run\_training} oversees the agent's training process; it initializes the appropriate agent, manages its training, and collects the resultant data. The \textit{run\_sweep} function manages hyperparameter sweeps. By systematically varying configurations, it seeks the optimal training setup. Each variation in the sweep leads to a new training session. Meanwhile, the \textit{run\_evaluation} function, although a placeholder in this context, is designed for post-training evaluation, testing the agent's adherence to certain performance metrics. Another significant aspect of the orchestrator is its integration with the wandb library. This ensures that crucial metrics, visualizations, and other data are continuously logged to the Weights \& Biases platform for subsequent analysis.

The \textbf{command-line interface (CLI)} serves as the interaction point between the user and the system's functionalities. It's built atop the argparse library, which facilitates the capture of command-line arguments. Users can clearly define their operational intentions by specifying modes \textit{(train, eval, or sweep)} and accompanying parameters such as alpha and agent\_type. Based on the provided arguments, the CLI determines the mode of operation: train leads to agent training, eval initiates the agent's evaluation, and sweep begins the process of hyperparameter optimization. 

\end{document}